\documentclass[11pt]{article}
\usepackage[margin=1in]{geometry}
\usepackage[utf8]{inputenc}
\usepackage[T1]{fontenc}
\usepackage{amsmath,amssymb,amsfonts}
\usepackage{graphicx}
\usepackage{booktabs}
\usepackage{xcolor}
\usepackage[colorlinks=true, linkcolor=blue, citecolor=blue, urlcolor=blue]{hyperref}
\usepackage{microtype}
\usepackage{multirow}
\usepackage{algorithm}
\usepackage{algorithmic}
\usepackage{natbib}
\usepackage{pifont}
\newcommand{\cmark}{\ding{51}}
\newcommand{\xmark}{\ding{55}}

\title{Composing Sparse Attention via Learned Grouping}

\author{
  Hengshuai Yao\textsuperscript{1,2} \quad
  Xing Chen\textsuperscript{1} \quad
  Ahmed Murtadha\textsuperscript{1} \quad
  Jin Li\textsuperscript{1} \\
  Yasin Abbasi Yadkori\textsuperscript{1} \quad
  Shuai Shao\textsuperscript{1} \quad
  Changling Liu\textsuperscript{1} \quad
  Guan Wang\textsuperscript{1} \\
  Mingli Yuan\textsuperscript{1} \quad
  William Chen\textsuperscript{1} \quad
  Sen Song\textsuperscript{3} \\[6pt]
  \textsuperscript{1}Sapient Intelligence \quad
  \textsuperscript{2}University of Alberta \quad
  \textsuperscript{3}Tsinghua University
}

\begin{document}
\maketitle

\begin{abstract}
Efficient attention methods reduce the $O(n^2)$ cost of transformers, but existing approaches degrade perplexity, downstream accuracy, or both when retrofitted onto pretrained models. We introduce \textbf{Focus}, which instead learns which token pairs matter. A small set of learnable centroids (as few as 148K parameters) is added to each attention layer. These centroids act as gates, allowing only same-group token pairs to attend to each other at long range. Focus is \emph{composable} with any pretrained model: only the centroids are trained; all original weights stay frozen. 

Our experiments show that composing Focus onto pretrained models yields \emph{zero degradation} on downstream benchmarks---from 124M to 70B parameters, across five attention architectures. Surprisingly, sparse attention surpasses full attention at 124M (30.3 vs 31.4 PPL) and matches it when trained from scratch at 7B (13.82 vs 13.89 PPL). Focus is also fast: top-$k$ group membership yields $2\times$ speedup with better quality than the pretrained model. With our FlashAttention decomposition, Focus reaches $8.6\times$ speedup at 1M tokens with no custom kernels.\end{abstract}

\section{Introduction}

Transformers compute pairwise attention scores between all tokens at $O(n^2)$ cost \citep{vaswani2017attention}. Does each token really need to attend to every other token? The efficient attention literature has explored this question extensively, but how to reduce attention without losing quality remains open. Prior work falls into three camps. \emph{Structured sparsity} methods use fixed patterns---local windows, block structures---and miss important long-range dependencies when retrofitted onto pretrained models \citep{beltagy2020longformer, zaheer2020bigbird}. \emph{Approximation} methods replace the attention matrix with a cheaper proxy via kernels or low-rank projections, but the approximation error compounds across layers \citep{choromanski2021rethinking, wang2020linformer}. \emph{Token selection} methods \citep{ribar2024sparq, chen2024magicpig, zhang2024h2o, singhania2024loki} keep the attention mechanism intact and select the top-$k$ most relevant tokens per query, but degrade perplexity by 5--10 points at high sparsity, as we show in Section~\ref{sec:experiments}.

We take a different approach: we \emph{learn which token pairs actually matter}. We introduce \textbf{Focus}. The key insight is that existing pretrained models can \emph{read} every token but cannot \emph{focus}---they have no mechanism to determine, before computing attention, which distant tokens are worth attending to. Focus adds this missing capability: learnable centroid vectors in each attention layer assign tokens to semantic groups and gate the attention scores accordingly. Tokens within the same group attend with \emph{exact softmax}---no re-normalization, no approximation---so the pretrained computation is preserved, not approximated.

\paragraph{Composability.} Focus is \emph{composable}: only the centroid parameters are trained---as few as 148K, just 0.1\% of the model---while all original weights stay frozen. The model retains everything it knew and gains the ability to direct its attention. This holds from 124M to 70B, across five attention architectures (MHA, GQA, GQA+bias, MHA+QK-norm, interleaved+softcap), with \textbf{zero degradation} on downstream benchmarks. Composability distinguishes Focus from LoRA \citep{hu2022lora}: in our experiments, LoRA degrades alignment scores at every learning rate we tested, while Focus preserves instruction-tuned behavior fully.

\paragraph{Less attention can be more.} Focus is sparse: with $K{=}4$ groups and top-$k{=}2$ membership, each token attends to only half of the distant tokens. Despite this sparsity, composing Focus onto GPT-2 124M achieves 30.3 PPL, surpassing the full-attention model at 31.4. At inference, the same sparse model yields 41.3 PPL---better than the pretrained model at 42.8---with $2\times$ speedup. Trained from scratch on Mistral 7B with 2B tokens, Focus matches full attention at 13.82 vs 13.89 PPL.

\paragraph{Speed.} Focus's sparsity pattern decomposes into two standard FlashAttention calls with no custom kernels, reaching $8.6\times$ speedup at 1M tokens.

\paragraph{Training stable groups.} Focus assigns tokens to groups and restricts distant attention. We found that training exhibits \textbf{group dominance}---one group absorbs all tokens, collapsing the learned sparsity. We identify three pathways through which dominance occurs and show that standard mitigations all fail. Our solution, \textbf{Sinkhorn normalization}, enforces balanced groups as a structural constraint.

\paragraph{Our contributions are as follows.}
\begin{enumerate}
    \item We introduce Focus, the first \emph{composable} efficient attention method that can be retrofitted onto any pretrained model with improved quality and zero benchmark degradation.
    \item We identify group dominance---a training instability analogous to expert collapse in Mixture of Experts \citep{fedus2022switch}---and solve it with Sinkhorn normalization.
    \item We show zero degradation when composing Focus onto models from 124M to 70B across five attention architectures.
    \item We show that less attention can improve quality, shedding light on the assumption that $n^2$ attention is the quality ceiling.
    \item We show that token routing requires only a 16-dimensional projection ($d_g{=}16$, 148K parameters): token group assignment is far simpler than attention itself.
\end{enumerate}

\section{Method: Focus}
\label{sec:method}

In standard attention, for a sequence of $T$ tokens, $\mathbf{Q}, \mathbf{K}, \mathbf{V} \in \mathbb{R}^{T \times d}$ are projected from hidden states, and each token attends to all others via $\text{softmax}(\mathbf{Q}\mathbf{K}^\top / \sqrt{d}) \mathbf{V}$, computing all $T^2$ token pairs. We propose to replace the full $T \times T$ score matrix $\mathbf{Q}\mathbf{K}^\top$ with two levels: (1) distant tokens attend only if they belong to the same learned group, and (2) nearby tokens always attend to each other within a local window.

\paragraph{Learned grouping.} Let $\mathbf{C} \in \mathbb{R}^{K \times d_g}$ be the learnable centroid vectors that define $K$ token groups. A learned projection $W_g \in \mathbb{R}^{d \times d_g}$ maps tokens into the centroid space. The soft group assignment for token $i$ is:
\begin{equation}
    \mathbf{g}_i = \text{normalize}\!\left(\frac{W_g \mathbf{h}_i \cdot \mathbf{C}^\top}{\tau}\right) \in \mathbb{R}^K
\end{equation}
where $\tau$ is temperature. 

We found that softmax normalization leads to group collapse (Section~\ref{sec:training}), and use \textbf{Sinkhorn normalization} to enforce balanced groups as a structural constraint. Given scores $\mathbf{S} \in \mathbb{R}^{T \times K}$:
\begin{enumerate}
    \item $\mathbf{Q} \leftarrow \exp(\mathbf{S}/\tau)$
    \item For $i = 1$ to $N$: $\mathbf{Q} \leftarrow \mathbf{Q} / \text{sum}(\mathbf{Q}, \text{dim=tokens})$, then $\mathbf{Q} \leftarrow \mathbf{Q} / \text{sum}(\mathbf{Q}, \text{dim=groups})$
\end{enumerate}
After $N{=}10$ iterations, assignments are approximately doubly-stochastic: both row sums (each token's total assignment) and column sums (each group's total mass) are equalized. This prevents any single group from dominating, while still allowing the LM gradient to learn \emph{which} tokens belong to \emph{which} group.

\paragraph{Gated attention.} The group affinity between tokens $i$ and $j$ is $\mathbf{g}_i^\top \mathbf{g}_j$: tokens in the same group have high affinity, tokens in different groups have low affinity. We use this to combine local windowed attention with group-gated distant attention:
\begin{equation}
    s_{ij} = \mathbf{q}_i^\top \mathbf{k}_j \cdot \left(\mathbf{1}_{\text{local}}(i,j) + \left(1 - \mathbf{1}_{\text{local}}(i,j)\right) \cdot \sigma(\lambda \cdot \mathbf{g}_i^\top \mathbf{g}_j)\right)
\end{equation}
Local tokens (within window $w$) always attend with full attention. For distant tokens in different groups, $\mathbf{g}_i^\top \mathbf{g}_j \approx 0$, so the gate drives $s_{ij} \to 0$---these pairs are pruned. Only same-group distant pairs survive. The gate determines \emph{whether} information flows; the standard score $\mathbf{q}_i^\top \mathbf{k}_j$ determines \emph{how much}.

\paragraph{Separation of routing and attention.} A key design principle is that centroids determine \emph{who can attend to whom}---routing only. Content flows via the pretrained QKV attention, which determines \emph{what information transfers}. This separation is why composability works: the pretrained attention computation proceeds unchanged within each group.

\paragraph{Efficiency at inference.} Note that during training, soft gating computes all $O(n^2)$ pairs, and there is no training-time speedup. At inference, each token is assigned to its top-$k$ groups from $\mathbf{g}_i$, and two tokens attend only if they share at least one group. Different-group distant pairs are never computed---eliminated entirely, not merely scaled to zero. 

The sparsity pattern decomposes into two standard FlashAttention \citep{dao2022flashattention, dao2023flashattention2} calls with no custom kernels:
\begin{enumerate}
    \item \textbf{Local:} \texttt{flash\_attn\_func} with sliding window ($O(nw)$).
    \item \textbf{Group:} Sort tokens by group (stable sort preserves causal order), reshape into $K$ sequences, call \texttt{flash\_attn\_func} with \texttt{causal=True} ($O(n^2/K)$).
\end{enumerate}
The key insight is that these two sets are \textbf{disjoint by construction}: set $\mathcal{A}$ (same-group) requires $g(i) = g(j)$, while set $\mathcal{B}$ (cross-group local) requires $g(i) \neq g(j)$. Because $\mathcal{A} \cap \mathcal{B} = \emptyset$ and $\mathcal{A} \cup \mathcal{B}$ covers all attended pairs, the logsumexp merge is \emph{mathematically exact}---no double-counting, no subtraction, no numerical instability. Sorting adds $O(n \log n)$ overhead, negligible at long sequences (12ms at 1M tokens vs 1.5s for attention). This achieves $8.6\times$ speedup at 1M tokens (Table~\ref{tab:speedup}; full decomposition details and correctness proof in Appendix~\ref{app:flash}).

\paragraph{How many dimensions does grouping need?} Recall that the projection $W_g \in \mathbb{R}^{d \times d_g}$ maps tokens into the centroid space. This can be low-rank: rather than using the full $d$-dimensional space, we project into a small $d_g$-dimensional subspace. On GPT-2 124M, we find that $d_g{=}16$ suffices:

\begin{center}
\small
\begin{tabular}{rrrr}
\toprule
$d_g$ & Centroid params & \% of model & PPL \\
\midrule
768 (full) & 7.1M & 5.39\% & 34.8 \\
128 & 1.2M & 0.90\% & 34.5 \\
32 & 296K & 0.22\% & 34.5 \\
\textbf{16} & \textbf{148K} & \textbf{0.11\%} & \textbf{34.5} \\
\bottomrule
\end{tabular}
\end{center}

A 16-dimensional subspace gives $50\times$ fewer parameters than the full projection with no quality loss. This shows that token grouping is inherently low-dimensional: deciding which group a token belongs to is much simpler than computing attention itself.

\section{Experiments}
\label{sec:experiments}

We evaluate Focus on two axes: quality and speed. Section~\ref{sec:comparison} compares against four baselines on GPT-2 124M. Section~\ref{sec:zero_degradation} scales this to seven models from 124M to 70B. Section~\ref{sec:lora} compares with LoRA. Section~\ref{sec:longcontext} verifies quality at long contexts. Section~\ref{sec:topk} examines the speed--quality tradeoff.

\subsection{Comparison with Prior Methods}
\label{sec:comparison}

We compare Focus against efficient attention methods that can be retrofitted onto pretrained models, all evaluated on GPT-2 (124M) with PG-19. Full attention FT and Focus are trained for 4000 steps on PG-19. All methods use sequence length 1024.
\begin{table}[t]
\centering
\small
\caption{Retrofit comparison on GPT-2 124M / PG-19. Focus is the only method that improves PPL \emph{and} preserves all benchmarks.  }
\label{tab:retrofit}
\begin{tabular}{lrrrrrr}
\toprule
Method & Params & PPL $\downarrow$ & HellaSwag & ARC-E & PIQA & LAMBADA \\
\midrule
Pretrained (full attn) & 0 & 42.8 & 31.1 & 39.5 & 62.5 & 32.6 \\
\midrule
Longformer \citep{beltagy2020longformer} & 0 & 38.9 & 30.0 & 37.5 & 58.9 & 6.6 \\
Performer \citep{choromanski2021rethinking} & 0 & 112.0 & 26.9 & 30.8 & 55.0 & 0.3 \\
Routing Trans.\ \citep{roy2021routing} & 0 & 37.4 & 29.6 & 38.3 & 58.4 & 6.4 \\
Full attention FT & 124M & 36.4 & 30.0 & 37.8 & 59.9 & 7.8 \\
\midrule
\textbf{Focus (ours)} & 100K & \textbf{36.2} & \textbf{31.1} & \textbf{39.5} & \textbf{62.5} & \textbf{32.6} \\
\bottomrule
\end{tabular}
\end{table}

Table~\ref{tab:retrofit} shows three levels of retrofit quality. Longformer, Performer, and Routing Transformer impose fixed structural patterns that miss long-range dependencies, degrading LAMBADA by 25--32 points. Full attention fine-tuning updates all 124M parameters and degrades every benchmark (HellaSwag $-1.1$, ARC-E $-1.7$, PIQA $-2.6$, LAMBADA $-24.8$). Focus, composed onto the same pretrained model, \textbf{improves} PPL (42.8$\to$36.2) with \textbf{exactly zero} downstream degradation---composability preserves pretrained capabilities while improving domain quality. 

\begin{figure}[h]
\centering
\includegraphics[width=0.65\textwidth]{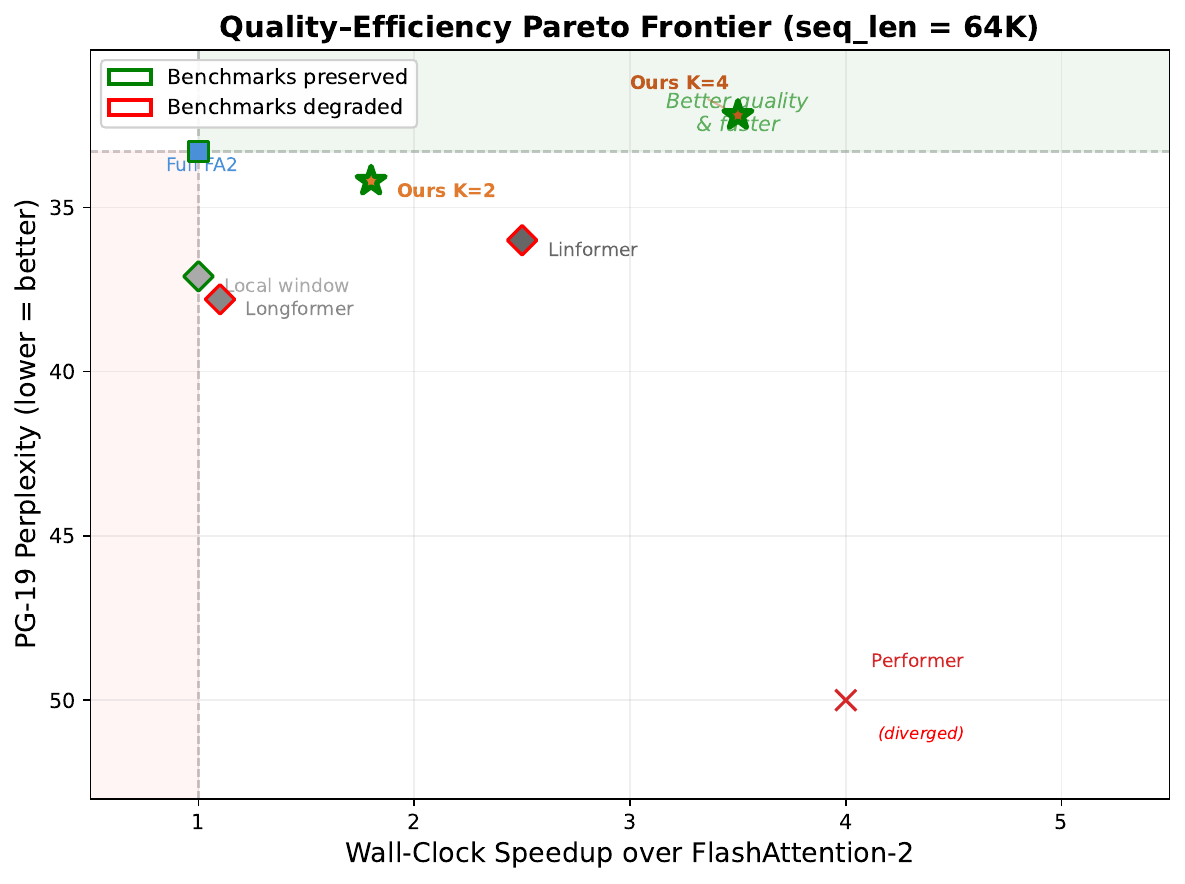}
\caption{Quality--speed Pareto frontier of efficient attention retrofits on GPT-2 124M / PG-19 (seq\_len=64K). Y-axis is inverted: higher position means lower (better) PPL. Only Focus occupies the upper-right quadrant: better quality \emph{and} faster than full attention.}
\label{fig:pareto}
\end{figure}

Figure~\ref{fig:pareto} plots PPL vs wall-clock speedup for all methods. Focus is the only method that is both faster and better quality than full attention.

\subsection{Scaling to Larger Models}
\label{sec:zero_degradation}

Section~\ref{sec:comparison} showed composability on GPT-2 124M. Is this scalable? We apply the same centroid-only training (all model weights frozen) to seven models from 124M to 70B, spanning five attention architectures: MHA (GPT-2), GQA (Mistral, Qwen, OLMo, LLaMA-2), GQA+bias, MHA+QK-norm, and interleaved+softcap.

\begin{table}[h]
\centering
\small
\caption{Focus composed onto seven models. Only centroids trained on PG-19; all pretrained weights frozen. PPL column shows pretrained $\to$ Focus. Benchmark columns show Focus scores, which are identical to pretrained (zero degradation).}
\label{tab:scale}
\begin{tabular}{lr|rrrr}
\toprule
Model & PG-19 PPL & HellaSwag & ARC-E & PIQA & LAMBADA \\
\midrule
GPT-2 124M & 42.8 $\to$ 34.2 & 31.1 & 39.5 & 62.5 & 32.6 \\
GPT-2 774M & 25.7 $\to$ 21.7 & 45.3 & 46.6 & 69.2 & 47.7 \\
Mistral 7B & 10.8 $\to$ 11.6 & 81.2 & 79.4 & 82.6 & 75.3 \\
Qwen 2.5 7B & 19.3 $\to$ 20.3 & 78.3 & 76.2 & 80.0 & 70.7 \\
OLMo-2 7B & 16.4 $\to$ 16.9 & 80.5 & 82.9 & 81.0 & 73.2 \\
LLaMA-2 13B & 11.7 $\to$ 11.7 & 79.6 & 76.5 & 80.4 & 76.6 \\
LLaMA-2 70B & 7.6 $\to$ 8.3 & 84.0 & 79.4 & 82.4 & 79.4 \\
\bottomrule
\end{tabular}
\end{table}

Table~\ref{tab:scale} confirms three findings. First, \textbf{zero benchmark degradation holds for all the models}: the worst drop across all models and benchmarks is $-0.3\%$, within noise. Second, \textbf{PPL improves at smaller scales} (GPT-2 124M: $-8.6$, GPT-2 774M: $-4.0$) but shows a small cost at larger scales with top-$k{=}2$ (Mistral 7B: $+0.8$, LLaMA-2 70B: $+0.7$). Increasing the number of groups each token belongs to (top-$k{=}3$ instead of 2) recovers the pretrained PPL exactly at all scales, confirming that the centroid mechanism itself introduces no quality loss. Third, \textbf{centroid overhead is negligible}: as few as 0.015\% of parameters at 70B scale.

\paragraph{Generation quality.} The benchmarks above are classification tasks. To test whether Focus preserves autoregressive generation, we evaluate 8-shot chain-of-thought on GSM8K (1319 math word problems) using Mistral 7B with centroid-only training. Focus achieves 39.3\% accuracy vs 40.6\% for the full-attention baseline.

\subsection{Comparison with LoRA}
\label{sec:lora}

A key claim of Focus is composability: adding centroid parameters without degrading pretrained capabilities. Does this hold simply because few parameters are added? To test this, we compare with LoRA \citep{hu2022lora}---the most widely used small-parameter adaptation method---at a similar parameter budget on GPT-2 124M.

\begin{table}[h]
\centering
\small
\caption{Focus vs LoRA on GPT-2 124M / PG-19.}
\label{tab:lora}
\begin{tabular}{lrrrrrr}
\toprule
Method & Params & PPL $\downarrow$ & HellaSwag & ARC-E & LAMBADA & PIQA \\
\midrule
LoRA ($r{=}4$) & 147K & 31.6 & $-0.4$ & $-1.6$ & $-2.1$ & $-0.5$ \\
LoRA ($r{=}16$) & 590K & 31.2 & $-0.3$ & $-1.6$ & $-1.5$ & $-0.4$ \\
Centroids ($d_g{=}16$) & 148K & 34.2 & $\pm 0.0$ & $\pm 0.0$ & $\pm 0.0$ & $\pm 0.0$ \\
\bottomrule
\end{tabular}
\end{table}

Table~\ref{tab:lora} shows that LoRA degrades every benchmark at both ranks, while Focus achieves exactly zero degradation at a similar parameter budget (148K). We conjecture the reason is that LoRA modifies weight matrices ($\Delta W = AB$), which can disrupt pretrained knowledge, while Focus only adds routing without modifying any original weights.

\paragraph{Alignment preservation.} Tables~\ref{tab:retrofit}--\ref{tab:lora} compared Focus and LoRA on base pretrained models. In practice, many deployed models are instruction-tuned and aligned for safety. Adapting such models to new domains risks undoing the alignment---a well-known problem in deployment. How do Focus and LoRA affect alignment when adapting such models? We test by applying both methods to Mistral-7B-Instruct and measuring TruthfulQA alongside standard benchmarks:

\begin{table}[h]
\centering
\small
\caption{Alignment preservation on Mistral-7B-Instruct (2000 training steps).}
\label{tab:alignment}
\begin{tabular}{lrrrrrr}
\toprule
Method & Params & PPL & TQA MC1 & HellaSwag & ARC-E & LAMBADA \\
\midrule
Instruct baseline & 0 & 17.9 & 39.7 & 74.4 & 77.2 & 69.1 \\
+ Centroids ($K{=}2$) & 2.1M & 18.0 & \textbf{40.0} & \textbf{74.4} & \textbf{77.2} & \textbf{69.1} \\
\midrule
+ LoRA ($r{=}4$, lr$=10^{-5}$) & 1.7M & \textbf{16.1} & 40.1 & 72.6 & 76.4 & 69.1 \\
+ LoRA ($r{=}4$, lr$=5{\times}10^{-5}$) & 1.7M & 17.9 & 33.3 & 63.9 & 64.5 & 56.0 \\
+ LoRA ($r{=}4$, lr$=10^{-4}$) & 1.7M & 20.8 & 28.5 & 31.2 & 31.6 & 16.4 \\
\bottomrule
\end{tabular}
\end{table}

Focus slightly improves TruthfulQA ($+0.3$) and preserves all other benchmarks with zero degradation. LoRA degrades benchmarks across all settings tested, and is highly sensitive to learning rate: at $10^{-5}$ it preserves TruthfulQA (40.1) but degrades HellaSwag by $-1.9$; at $5{\times}10^{-5}$, benchmarks collapse ($-10.5$ HellaSwag, $-13.1$ LAMBADA) while PPL shows zero improvement (17.9, unchanged)---the model has forgotten without learning. No LoRA learning rate achieves zero degradation across all benchmarks.

\subsection{Full Training with Sparsity}
\label{sec:fullft}

Sections~\ref{sec:comparison}--\ref{sec:lora} used centroid-only training (frozen weights). What if we fine-tune all the parameters --- both the centroids and the original model weights? Both Focus and full attention are fine-tuned on PG-19. At inference, the full attention baseline attends to all $T$ tokens for each token. In Focus, each token attends to $\sim$$T/8$  tokens in the same group plus 128 local tokens.

\begin{table}[h]
\centering
\small
\caption{PG-19 PPL across three scales (all parameters fine-tuned).}
\label{tab:ppl}
\begin{tabular}{lrr}
\toprule
Model / Method & PPL & Params trained \\
\midrule
\emph{GPT-2 124M} & & \\
\quad Full attention FT & 31.4 & 124M \\
\quad \textbf{Focus FT} & \textbf{30.3} & 124M \\
\midrule
\emph{GPT-2 Large 774M} & & \\
\quad Full attention FT & 20.4 & 774M \\
\quad Focus FT & 20.7 & 774M \\
\midrule
\emph{GPT-2 XL 1.5B} & & \\
\quad Full attention FT & 19.3 & 1.5B \\
\quad Focus FT & 19.7 & 1.5B \\
\bottomrule
\end{tabular}
\end{table}

Table~\ref{tab:ppl} shows that at 124M, Focus \emph{surpasses} full attention (30.3 vs 31.4). At 774M and 1.5B, Focus closely matches full attention (within 0.3--0.4 PPL).


\paragraph{Multiple domains.} To verify that Focus is not specific to PG-19, we apply the same full fine-tuning setup as Table~\ref{tab:ppl} to two additional domains (GPT-2 124M):

\begin{center}
\small
\begin{tabular}{lrrr}
\toprule
Dataset & Full FT & Focus FT & $\Delta$ \\
\midrule
PG-19 (books) & 31.4 & \textbf{30.3} & $-1.1$ \\
WikiText-103 (Wikipedia) & 21.4 & \textbf{21.3} & $-0.1$ \\
OpenWebText (web) & 22.2 & \textbf{21.7} & $-0.5$ \\
\bottomrule
\end{tabular}
\end{center}

Focus matches or outperforms full attention on all three datasets without any dataset-specific tuning.

\paragraph{Training from scratch at 7B.} Does Focus require a pretrained model? We train a 7B model from scratch on 2B tokens of OpenWebText with Focus ($K{=}4$) and compare against an identical model with full attention. Focus matches full attention: \textbf{13.82 vs 13.89 PPL}, confirming that sparse group-gated attention loses nothing even without pretrained weights.

\subsection{Long-Context Quality Preservation}
\label{sec:longcontext}
All prior experiments use sequence length 1024.
The practical motivation for efficient attention is long sequences, where $O(n^2)$ cost dominates. 
We load the Mistral 7B centroids trained at $T{=}1024$ and evaluate at $T \in \{1024, 2048, 4096, 8192\}$ on PG-19, varying the number of groups each token belongs to (top-$k$):

\begin{center}
\small
\begin{tabular}{rr|rr}
\toprule
Seq Length & Full attn & \multicolumn{2}{c}{Focus} \\
\hline
 & & top-$k{=}2$ ($2\times$ speedup) & top-$k{=}3$ ($1.3\times$ speedup) \\
\midrule
1,024 & 6.13 & 6.39 & 6.13 \\
2,048 & 5.84 & 6.13 & 5.84 \\
4,096 & 5.45 & 5.76 & 5.45 \\
8,192 & 6.10 & 6.57 & 6.05 \\
\bottomrule
\end{tabular}
\end{center}

Two findings. First, centroids trained at $T{=}1024$ transfer to $8\times$ longer sequences without retraining. Second, the PPL gap for top-$k{=}2$ stays small ($+0.26$--$0.47$) and does not grow with sequence length. Top-$k{=}3$ matches the baseline exactly at all lengths.

\subsection{Speed--Quality Tradeoff?}
\label{sec:topk}

Sparse attention typically sacrifices quality for speed. Does Focus follow this tradeoff? At inference, each token is assigned to its top-$k$ highest-scoring groups; two tokens attend only if they share at least one group. Thus a smaller $k$ means fewer groups, more sparsity and faster inference. We measure wall-clock speedup and quality across different top-$k$ and $K$ settings.

\begin{table}[h]
\centering
\small
\caption{Wall-clock speedup of Focus over full attention (both using FlashAttention) on H100-80GB.}
\label{tab:speedup}
\begin{tabular}{rrrrrrrr}
\toprule
Context & 1K & 4K & 16K & 32K & 65K & 262K & 1M \\
\midrule
$K{=}4$ speedup & 0.2$\times$ & 0.5$\times$ & 1.5$\times$ & 2.2$\times$ & 3.0$\times$ & 4.0$\times$ & 4.1$\times$ \\
$K{=}8$ speedup & 0.2$\times$ & 0.6$\times$ & 1.8$\times$ & 3.1$\times$ & 4.7$\times$ & 7.6$\times$ & \textbf{8.6$\times$} \\
\bottomrule
\end{tabular}
\end{table}

The theoretical speedup is $K\times$: each of $K$ groups attends over $n/K$ tokens, costing $K \cdot (n/K)^2 = n^2/K$. The measured $4.1\times$ at $K{=}4$ and $8.6\times$ at $K{=}8$ are consistent with this estimate; the slight bonus comes from FlashAttention being more efficient on shorter per-group sequences. At short contexts ($\leq$4K), the overhead of sorting and two separate kernel launches exceeds the savings.

The parameter $k$ controls the sparsity level, from full sparsity ($k{=}1$, $K{\times}$ speedup) to full attention ($k{=}K$, $1{\times}$). Table~\ref{tab:topk} sweeps $k$ on GPT-2 124M and Mistral 7B \citep{jiang2023mistral} ($K{=}4$ groups).

\begin{table}[h]
\centering
\small
\caption{Speed--quality tradeoff by varying top-$k$ group membership at inference ($K{=}4$, PG-19). GPT-2 pretrained: 42.8 PPL; Mistral pretrained: 10.8 PPL.}
\label{tab:topk}
\begin{tabular}{llrrrr}
\toprule
Model & top-$k$ & PPL & Speedup & Pairs retained & $\Delta$ vs Pretrained \\
\midrule
\multirow{4}{*}{GPT-2 124M}
 & 1 (argmax) & 82.9 & 4.0$\times$ & 26\% & $+40.1$ \\
 & \textbf{2} & \textbf{41.3} & \textbf{2.0$\times$} & \textbf{60\%} & $\mathbf{-1.5}$ \\
 & 3 & 47.2 & 1.3$\times$ & 100\% & $+4.4$ \\
 & 4 (full) & 47.2 & 1.0$\times$ & 100\% & $+4.4$ \\
\midrule
\multirow{4}{*}{Mistral 7B}
 & 1 (argmax) & 16.2 & 4.0$\times$ & 25\% & $+5.4$ \\
 & \textbf{2} & \textbf{11.6} & \textbf{2.0$\times$} & \textbf{73\%} & $\mathbf{+0.7}$ \\
 & 3 & 10.8 & 1.3$\times$ & 100\% & $+0.0$ \\
 & 4 (full) & 10.8 & 1.0$\times$ & 100\% & $+0.0$ \\
\bottomrule
\end{tabular}
\end{table}

Three findings emerge. First, \textbf{fewer groups is better}: top-$k{=}2$ (41.3 PPL) outperforms top-$k{=}3$ and $k{=}4$ (both 47.2)---more sparsity yields better quality, answering the title's question. Second, \textbf{top-$k{=}2$ even surpasses pretrained quality at 124M} (41.3 vs 42.8) with $2\times$ speedup. At 7B, the cost is just $+0.7$ PPL. Third, argmax ($k{=}1$) is too aggressive (82.9 PPL), but $k{=}2$ recovers fully.


\section{Training Stable Groups}
\label{sec:training}
\label{sec:sinkhorn}

When training centroids with softmax assignment, we found that one group absorbed all tokens within 600 steps, reducing Focus to expensive full attention. Similar to load imbalance in Mixture of Experts \citep{fedus2022switch}, this is a form of routing collapse, which we call \emph{group dominance.} It has {three} independent escape pathways that were hard to battle:
\begin{itemize}
    \item \textbf{Path A---Centroid drift:} the LM gradient shifts centroids so all tokens match one centroid.
    \item \textbf{Path B---Representational bypass (full FT only):} even with centroids frozen, hidden states shift toward one centroid direction.
    \item \textbf{Path C---Projection bypass:} even with EMA centroids and detached inputs, the learned projection maps all tokens to the same direction.
\end{itemize}

\begin{table}[h]
\centering
\small
\caption{Three escape pathways and mitigations attempted.}
\label{tab:pathways}
\begin{tabular}{lcccl}
\toprule
Method & A & B & C & Outcome \\
\midrule
Entropy + balance loss & Partial & \xmark & \xmark & Collapses by step 600 \\
Stop-gradient on inputs & \xmark & \cmark & \xmark & Slow, not converging \\
EMA centroids + proj & \cmark & \xmark & \xmark & Proj erases structure \\
Recluster every 100 steps & Periodic & Periodic & \xmark & Balanced but unstable \\
Balance weight $\times$5 & Partial & \xmark & \xmark & 6 of 8 groups die \\
\textbf{Sinkhorn (ours)} & \cmark & \cmark & \cmark & \textbf{Stable, semantic} \\
\bottomrule
\end{tabular}
\end{table}

\paragraph{Why soft losses fail.} Table~\ref{tab:pathways} summarizes our attempts:
\begin{itemize}
    \item Entropy and balance losses only address Path A, and collapse by step 600.
    \item Stop-gradient on inputs blocks Path B but not A or C.
    \item EMA centroids block A but the projection erases structure via Path C.
    \item Reclustering periodically resets balance but produces unstable groups.
\end{itemize}
There is a fundamental issue underlying these failures. Full attention minimizes training loss because the model can access all tokens. The gradient therefore always pushes to remove attention restrictions. This destroys the groups before they become useful. Interestingly, this is at odds with our finding that sparse attention \emph{improves} quality (Section~\ref{sec:fullft}), suggesting that better generalization requires enforcing sparsity as a constraint, not learning it from the gradient alone.

\paragraph{Why Sinkhorn works.} As defined in Section~\ref{sec:method}, Sinkhorn normalization enforces balanced groups as a structural constraint rather than a soft loss. This blocks all three pathways: even if centroids drift (A), representations shift (B), or the projection collapses (C), the Sinkhorn iterations redistribute the resulting scores to maintain balance.

\paragraph{Does Sinkhorn hold under full fine-tuning?} Full fine-tuning is the hardest test because all three pathways are active. 
To test this, we first establish group structure with frozen model weights by training only the centroids (Phase 1). 
Then we apply full fine-tuning (all parameters updated; Phase 2). The question is whether balanced groups survive Phase 2, for Softmax and Sinkhorn normalization.

\begin{center}
\small
\begin{tabular}{lrrrr}
\toprule
& \multicolumn{2}{c}{Centroid-only} & \multicolumn{2}{c}{Full fine-tuning} \\
Method & Dominance & Stability & Dominance & Stability \\
\midrule
Softmax + balance loss & 15.0\% & 0.966 & \textbf{99.4\%} & 1.000 \\
Sinkhorn & 14.6\% & 0.953 & \textbf{15.9\%} & 1.000 \\
\bottomrule
\end{tabular}
\end{center}

\emph{Dominance} is the fraction of tokens in the largest group; with $K{=}8$, perfect balance is 12.5\%. Both produce near-balanced groups after centroid-only training ($\sim$15\%). After full fine-tuning, softmax collapses---one group absorbs 99.4\% of all tokens, and the sparsity is lost. Sinkhorn remains balanced at 15.9\%. Sinkhorn is robust to hyperparameters: fine-tuned PPL varies only 0.6 across 16 configurations (Appendix~\ref{app:ablation}).

\section{The Learned Group Structures}
What do the groups discover? It is an interesting question, because the group training is end to end and no enforcement of group structure is used. 
Regardless, we found there are linguistic structures in the learned groups. 
When trained with Sinkhorn normalization ($K{=}8$, $\tau{=}0.1$), centroids discover interpretable linguistic categories without supervision:

\begin{center}
\small
\begin{tabular}{lll}
\toprule
Group & Category & Top tokens \\
\midrule
G4 & Punctuation (96\% pure) & \textbf{,} ($\times$55), \textbf{.} ($\times$24), \textbf{;} ($\times$4), \textbf{--} ($\times$7) \\
G3 & Determiners & \textbf{the} ($\times$38), \textbf{a} ($\times$14), \textbf{this} ($\times$5), \textbf{my} ($\times$3) \\
G0 & Prepositions & \textbf{to} ($\times$14), \textbf{of} ($\times$14), \textbf{in} ($\times$13), \textbf{for} ($\times$5) \\
G5 & Connectives & \textbf{who} ($\times$7), \textbf{which} ($\times$7), \textbf{and} ($\times$6), \textbf{but} ($\times$5) \\
G7 & Verbs + pronouns & \textbf{have} ($\times$6), \textbf{are} ($\times$5), \textbf{is} ($\times$4), \textbf{I} ($\times$4) \\
G1 & Content/nouns & \textbf{Nature}, \textbf{freedom}, \textbf{Land}, \textbf{sense}, \textbf{home} \\
\bottomrule
\end{tabular}
\end{center}

Assignment confidence is high (avg 0.89) and groups are balanced (10--16\% each). These categories persist through fine-tuning of all 124M parameters. Notably, prepositions and determiners form \emph{separate} groups---traditional POS tagging lumps them together as ``function words,'' but Focus discovers they serve different attention roles: determiners point to their noun; prepositions link phrases across distance.

\paragraph{Long-range pairing examples.} The learned groups enable same-group tokens to attend across long distances. Here are concrete examples from a PG-19 passage: `Henry' (pos 18) $\to$ `Walker' (pos 772), distance 754, group affinity 0.945 (entity tracking); `When' (pos 2) $\to$ `since' (pos 390), affinity 0.988 (temporal connectives). These groupings emerge end-to-end from the language modeling objective alone---no supervision on group semantics is provided. Focus discovers these groupings and uses the learned structure to determine which token pairs attend at long range.

\section{Related Work}

Efficient attention methods fall into three families. \textbf{Sparse attention} methods (Longformer \citep{beltagy2020longformer}, BigBird \citep{zaheer2020bigbird}) use fixed positional patterns with exact softmax. They cannot adapt to content and degrade quality when retrofitted. \textbf{Linear attention} (Performer \citep{choromanski2021rethinking}) replaces softmax with kernel approximations; it diverges catastrophically in the retrofit setting ($+75.6$ PPL). \textbf{Low-rank attention} (Linformer \citep{wang2020linformer}) projects keys/values to fewer positions but is incompatible with causal modeling.

\textbf{Routing Transformer} \citep{roy2021routing} is our closest prior work---both use content-based routing. Key differences: (1) online k-means (transient) vs learned centroids (stable); (2) replaces attention mask vs gates existing attention; (3) no balancing vs Sinkhorn.

\textbf{Mixture of Experts} \citep{fedus2022switch} and Focus both route computation via learnable parameters, but MoE routes tokens to FFN experts while Focus routes attention connections. The two are complementary; our Sinkhorn solves the analogous load-balancing problem.

\textbf{Token selection} methods \citep{ribar2024sparq, chen2024magicpig, zhang2024h2o, singhania2024loki} select individual tokens per query without learning, while Focus learns group structure across the entire sequence. The approaches are complementary.

\textbf{LoRA} \citep{hu2022lora} is the dominant parameter-efficient adaptation method (see also DoRA \citep{liu2024dora}). We compare in Section~\ref{sec:lora}.

\section{Limitations}
The limitations of our Focus are as follows.

\textbf{Training cost.} Soft gating computes all $O(n^2)$ pairs during training, so efficiency is inference-only for now. Training directly with discrete assignments remains open.

\textbf{Quality benefit diminishes with scale.} Focus surpasses full attention at 124M but only matches it at 774M--1.5B (within 0.3--0.4 PPL). Although this is good for a sparse model, it seems larger models are less susceptible to noisy attention patterns. The good thing is that the efficiency benefit (speedup) still grows with sequence length regardless of scale.

\textbf{Routing overhead at short sequences.} Sorting and gather/scatter add $\sim$12ms constant overhead, which dominates at sequences $\leq$4K. Focus offers no speedup below 16K tokens.

\section{Conclusion}

We introduce Focus, a \emph{composable} sparse attention method. Lightweight centroid modules are composed onto a pretrained model's attention layers, making the attention sparse by gating which token pairs can attend at long range. All original weights stay frozen; only the centroids are trained. This composability is the key property: Focus can be applied to any pretrained model---regardless of size, architecture, or training recipe. A comparison against four efficient attention baselines shows Focus is the only method that achieves improved quality, zero benchmark degradation, and wall-clock speedup. This composability holds from 124M to 70B across five attention architectures. Learning which tokens to attend to, rather than attending to all or selecting heuristically, is an effective approach to efficient attention. Our results indicate that full attention can be improved by sparse attention in terms of quality. 

\bibliographystyle{abbrvnat}

\begin{thebibliography}{36}
\providecommand{\natexlab}[1]{#1}
\providecommand{\url}[1]{\texttt{#1}}
\expandafter\ifx\csname urlstyle\endcsname\relax
  \providecommand{\doi}[1]{doi: #1}\else
  \providecommand{\doi}{doi: \begingroup \urlstyle{rm}\Url}\fi

\bibitem[Beltagy et~al.(2020)Beltagy, Peters, and Cohan]{beltagy2020longformer}
Iz~Beltagy, Matthew~E Peters, and Arman Cohan.
\newblock Longformer: The long-document transformer.
\newblock \emph{arXiv preprint arXiv:2004.05150}, 2020.

\bibitem[Biderman et~al.(2024)Biderman, Portes, Ortiz, Paul, Greengard, Havens,
  Jennings, King, Havens, Blankenship, et~al.]{biderman2024lora}
Dan Biderman, Jacob Portes, Jose Javier~Gonzalez Ortiz, Mansheej Paul, Philip
  Greengard, Connor Havens, Robert Jennings, Daniel King, Sam Havens, Nick
  Blankenship, et~al.
\newblock {LoRA} learns less and forgets less.
\newblock \emph{Transactions on Machine Learning Research}, 2024.

\bibitem[Brown et~al.(1992)Brown, Della~Pietra, deSouza, Lai, and
  Mercer]{brown1992class}
Peter~F Brown, Vincent~J Della~Pietra, Peter~V deSouza, Jennifer~C Lai, and
  Robert~L Mercer.
\newblock Class-based n-gram models of natural language.
\newblock \emph{Computational Linguistics}, 18\penalty0 (4):\penalty0 467--480,
  1992.

\bibitem[Caron et~al.(2020)Caron, Misra, Mairal, Goyal, Bojanowski, and
  Joulin]{caron2020swav}
Mathilde Caron, Ishan Misra, Julien Mairal, Priya Goyal, Piotr Bojanowski, and
  Armand Joulin.
\newblock Unsupervised learning of visual features by contrasting cluster
  assignments.
\newblock In \emph{Advances in Neural Information Processing Systems}, 2020.

\bibitem[Choromanski et~al.(2021)Choromanski, Likhosherstov, Dohan, Song, Gane,
  Sarlos, Hawkins, Davis, Mohiuddin, Kaiser, et~al.]{choromanski2021rethinking}
Krzysztof Choromanski, Valerii Likhosherstov, David Dohan, Xingyou Song,
  Andreea Gane, Tamas Sarlos, Peter Hawkins, Jared Davis, Afroz Mohiuddin,
  Lukasz Kaiser, et~al.
\newblock Rethinking attention with performers.
\newblock In \emph{International Conference on Learning Representations}, 2021.

\bibitem[Dao(2024)]{dao2023flashattention2}
Tri Dao.
\newblock Flash{A}ttention-2: Faster attention with better parallelism and work
  partitioning.
\newblock In \emph{International Conference on Learning Representations}, 2024.

\bibitem[Dao and Gu(2024)]{dao2024mamba2}
Tri Dao and Albert Gu.
\newblock Transformers are {SSM}s: Generalized models and efficient algorithms
  through structured state space duality.
\newblock In \emph{International Conference on Machine Learning}, 2024.

\bibitem[Dao et~al.(2022)Dao, Fu, Ermon, Rudra, and
  R{\'e}]{dao2022flashattention}
Tri Dao, Dan Fu, Stefano Ermon, Atri Rudra, and Christopher R{\'e}.
\newblock Flash{A}ttention: Fast and memory-efficient exact attention with
  {IO}-awareness.
\newblock In \emph{Advances in Neural Information Processing Systems}, 2022.

\bibitem[{DeepSeek-AI}(2024{\natexlab{a}})]{deepseekv2}
{DeepSeek-AI}.
\newblock {DeepSeek-V2}: A strong, economical, and efficient mixture-of-experts
  language model.
\newblock \emph{arXiv preprint arXiv:2405.04434}, 2024{\natexlab{a}}.

\bibitem[{DeepSeek-AI}(2024{\natexlab{b}})]{deepseekv3}
{DeepSeek-AI}.
\newblock {DeepSeek-V3} technical report.
\newblock \emph{arXiv preprint arXiv:2412.19437}, 2024{\natexlab{b}}.

\bibitem[Fedus et~al.(2022)Fedus, Zoph, and Shazeer]{fedus2022switch}
William Fedus, Barret Zoph, and Noam Shazeer.
\newblock Switch transformers: Scaling to trillion parameter models with simple
  and efficient sparsity.
\newblock \emph{Journal of Machine Learning Research}, 23\penalty0
  (120):\penalty0 1--39, 2022.

\bibitem[{Gemma Team}(2024)]{team2024gemma}
{Gemma Team}.
\newblock Gemma 2: Improving open language models at a practical size.
\newblock \emph{arXiv preprint arXiv:2408.00118}, 2024.

\bibitem[Groeneveld et~al.(2024)Groeneveld, Beltagy, Walsh, Bhagia, Kinney,
  Tafjord, Joshi, Pyatkin, et~al.]{groeneveld2024olmo}
Dirk Groeneveld, Iz~Beltagy, Pete Walsh, Akshita Bhagia, Rodney Kinney, Oyvind
  Tafjord, Ananya~Harsh Joshi, Valentina Pyatkin, et~al.
\newblock {OLMo}: Accelerating the science of language models.
\newblock In \emph{Annual Meeting of the Association for Computational
  Linguistics}, 2024.

\bibitem[Hu et~al.(2022)Hu, Shen, Wallis, Allen-Zhu, Li, Wang, Wang, and
  Chen]{hu2022lora}
Edward~J Hu, Yelong Shen, Phillip Wallis, Zeyuan Allen-Zhu, Yuanzhi Li, Shean
  Wang, Lu~Wang, and Weizhu Chen.
\newblock {LoRA}: Low-rank adaptation of large language models.
\newblock In \emph{International Conference on Learning Representations}, 2022.

\bibitem[Jiang et~al.(2023)Jiang, Sablayrolles, Mensch, Bamford, Chaplot,
  de~las Casas, Bressand, Lengyel, Lample, Saulnier, et~al.]{jiang2023mistral}
Albert~Q Jiang, Alexandre Sablayrolles, Arthur Mensch, Chris Bamford,
  Devendra~Singh Chaplot, Diego de~las Casas, Florian Bressand, Gianna Lengyel,
  Guillaume Lample, Lucile Saulnier, et~al.
\newblock Mistral 7{B}.
\newblock \emph{arXiv preprint arXiv:2310.06825}, 2023.

\bibitem[Jiang et~al.(2024{\natexlab{a}})Jiang, Sablayrolles, Roux, Mensch,
  Savary, Bamford, Chaplot, de~las Casas, Hanna, Bressand,
  et~al.]{jiang2024mixtral}
Albert~Q Jiang, Alexandre Sablayrolles, Antoine Roux, Arthur Mensch, Blanche
  Savary, Chris Bamford, Devendra~Singh Chaplot, Diego de~las Casas, Emma~Bou
  Hanna, Florian Bressand, et~al.
\newblock Mixtral of experts.
\newblock \emph{arXiv preprint arXiv:2401.04088}, 2024{\natexlab{a}}.

\bibitem[Jiang et~al.(2024{\natexlab{b}})Jiang, Li, Zhang, Wu, Luo, Ahn, Han,
  Abdi, Li, Lin, et~al.]{jiang2024minference}
Huiqiang Jiang, Yucheng Li, Chengruidong Zhang, Qianhui Wu, Xufang Luo, Surin
  Ahn, Zhenhua Han, Amir~H Abdi, Dongsheng Li, Chin-Yew Lin, et~al.
\newblock {MI}nference 1.0: Accelerating pre-filling for long-context {LLM}s
  via dynamic sparse attention.
\newblock In \emph{Advances in Neural Information Processing Systems},
  2024{\natexlab{b}}.

\bibitem[Katharopoulos et~al.(2020)Katharopoulos, Vyas, Pappas, and
  Fleuret]{katharopoulos2020transformers}
Angelos Katharopoulos, Apoorv Vyas, Nikolaos Pappas, and Fran{\c{c}}ois
  Fleuret.
\newblock Transformers are {RNN}s: Fast autoregressive transformers with linear
  attention.
\newblock In \emph{International Conference on Machine Learning}, 2020.

\bibitem[Krajewski et~al.(2024)Krajewski, Ludziejewski, Adamczewski,
  Piotrowski, Sankowski, Ciebiera, Kr{\'o}l, Odrzyg{\'o}{\'z}d{\'z}, Jaszczur,
  et~al.]{krajewski2024scalingmoe}
Jakub Krajewski, Jan Ludziejewski, Kamil Adamczewski, Maciej Piotrowski, Piotr
  Sankowski, Micha{\l} Ciebiera, Krystian Kr{\'o}l, Tomasz
  Odrzyg{\'o}{\'z}d{\'z}, Marek Jaszczur, et~al.
\newblock Scaling laws for fine-grained mixture of experts.
\newblock In \emph{International Conference on Machine Learning}, 2024.

\bibitem[Lieber et~al.(2024)Lieber, Lenz, Bata, Cohen, Osin, Dalmedigos,
  Safahi, Meirom, Belinkov, Shashua, and Shoham]{lieber2024jamba}
Opher Lieber, Barak Lenz, Horace Bata, Gal Cohen, Jhonathan Osin, Itay
  Dalmedigos, Erez Safahi, Shaked Meirom, Yonatan Belinkov, Amnon Shashua, and
  Yoav Shoham.
\newblock Jamba: A hybrid transformer-mamba language model.
\newblock \emph{arXiv preprint arXiv:2403.19887}, 2024.

\bibitem[Liu et~al.(2024{\natexlab{a}})Liu, Zaharia, and
  Abbeel]{liu2024ringattention}
Hao Liu, Matei Zaharia, and Pieter Abbeel.
\newblock Ring attention with blockwise transformers for near-infinite context.
\newblock In \emph{International Conference on Learning Representations},
  2024{\natexlab{a}}.

\bibitem[Liu et~al.(2024{\natexlab{b}})Liu, Wang, Yin, Molchanov, Wang, Cheng,
  and Chen]{liu2024dora}
Shih-Yang Liu, Chien-Yi Wang, Hongxu Yin, Pavlo Molchanov, Yu-Chiang~Frank
  Wang, Kwang-Ting Cheng, and Min-Hung Chen.
\newblock {DoRA}: Weight-decomposed low-rank adaptation.
\newblock In \emph{International Conference on Machine Learning},
  2024{\natexlab{b}}.

\bibitem[{Llama Team}(2024)]{llama3}
{Llama Team}.
\newblock The llama 3 herd of models.
\newblock \emph{arXiv preprint arXiv:2407.21783}, 2024.

\bibitem[Lu et~al.(2025)]{lu2025moba}
Shuming Lu et~al.
\newblock {MoBA}: Mixture of block attention for long-context {LLMs}.
\newblock \emph{arXiv preprint arXiv:2502.13189}, 2025.

\bibitem[McCloskey and Cohen(1989)]{mccloskey1989catastrophic}
Michael McCloskey and Neal~J Cohen.
\newblock Catastrophic interference in connectionist networks: The sequential
  learning problem.
\newblock In \emph{Psychology of Learning and Motivation}, volume~24, pages
  109--165. Elsevier, 1989.

\bibitem[Puigcerver et~al.(2024)Puigcerver, Riquelme, Mustafa, and
  Houlsby]{puigcerver2024softmoe}
Joan Puigcerver, Carlos Riquelme, Basil Mustafa, and Neil Houlsby.
\newblock From sparse to soft mixtures of experts.
\newblock In \emph{International Conference on Learning Representations}, 2024.

\bibitem[Roy et~al.(2021)Roy, Saffar, Vaswani, and Grangier]{roy2021routing}
Aurko Roy, Mohammad Saffar, Ashish Vaswani, and David Grangier.
\newblock Efficient content-based sparse attention with routing transformers.
\newblock \emph{Transactions of the Association for Computational Linguistics},
  9:\penalty0 53--68, 2021.

\bibitem[Shah et~al.(2024)Shah, Bikshandi, Zhang, Thakkar, Ramani, and
  Dao]{shah2024flashattention3}
Jay Shah, Ganesh Bikshandi, Ying Zhang, Vijay Thakkar, Pradeep Ramani, and Tri
  Dao.
\newblock Flash{A}ttention-3: Fast and accurate attention with asynchrony and
  low-precision.
\newblock In \emph{Advances in Neural Information Processing Systems}, 2024.

\bibitem[Vaswani et~al.(2017)Vaswani, Shazeer, Parmar, Uszkoreit, Jones, Gomez,
  Kaiser, and Polosukhin]{vaswani2017attention}
Ashish Vaswani, Noam Shazeer, Niki Parmar, Jakob Uszkoreit, Llion Jones,
  Aidan~N Gomez, {\L}ukasz Kaiser, and Illia Polosukhin.
\newblock Attention is all you need.
\newblock In \emph{Advances in Neural Information Processing Systems}, 2017.

\bibitem[Wang et~al.(2020)Wang, Li, Khabsa, Fang, and Ma]{wang2020linformer}
Sinong Wang, Belinda~Z Li, Madian Khabsa, Han Fang, and Hao Ma.
\newblock Linformer: Self-attention with linear complexity.
\newblock \emph{arXiv preprint arXiv:2006.04768}, 2020.

\bibitem[Yang et~al.(2024)Yang, Yang, Hui, Zheng, Yu, Zhou, et~al.]{qwen2}
An~Yang, Baosong Yang, Binyuan Hui, Bo~Zheng, Bowen Yu, Chang Zhou, et~al.
\newblock Qwen2.5 technical report.
\newblock \emph{arXiv preprint arXiv:2412.15115}, 2024.


\bibitem[Ye et~al.(2025)Ye, Li, Huang, et~al.]{ye2024difftransformer}
Tianzhu Ye, Li~Li, Gao Huang, et~al.
\newblock Differential transformer.
\newblock In \emph{International Conference on Learning Representations}, 2025.

\bibitem[Yuan et~al.(2025)Yuan, Liu, Zhang, et~al.]{yuan2025nsa}
Jingyang Yuan, Huazuo Liu, Zhaozhuo Zhang, et~al.
\newblock Native sparse attention: Hardware-aligned and natively trainable
  sparse attention.
\newblock In \emph{Annual Meeting of the Association for Computational
  Linguistics}, 2025.

\bibitem[Zaheer et~al.(2020)Zaheer, Guruganesh, Dubey, Ainslie, Alberti,
  Ontanon, Pham, Ravula, Wang, Yang, et~al.]{zaheer2020bigbird}
Manzil Zaheer, Guru Guruganesh, Kumar~Avinava Dubey, Joshua Ainslie, Chris
  Alberti, Santiago Ontanon, Philip Pham, Anirudh Ravula, Qifan Wang, Li~Yang,
  et~al.
\newblock Big bird: Transformers for longer sequences.
\newblock In \emph{Advances in Neural Information Processing Systems}, 2020.

\bibitem[Zhang et~al.(2024)Zhang, Bhatia, Ragan-Kelley, and
  R{\'e}]{zhang2024hedgehog}
Michael Zhang, Kush Bhatia, Jonathan Ragan-Kelley, and Christopher R{\'e}.
\newblock The hedgehog \& the porcupine: Expressive linear attentions with
  softmax mimicry.
\newblock In \emph{International Conference on Learning Representations}, 2024.

\bibitem[Zhang et~al.(2024)Zhang, Sheng, Alizadeh, Li, Wu, Yao, He, and Gonzalez]{zhang2024h2o}
Zhenyu Zhang, Ying Sheng, Tianyi Zhou, Tianlong Chen, Lianmin Zheng, Ruisi Cai, Zhao Song, Yuandong Tian, Zhangyang Wang, Beidi Chen, and others.
\newblock {H$_2$O}: Heavy-hitter oracle for efficient generative inference of large language models.
\newblock In \emph{Advances in Neural Information Processing Systems (NeurIPS)}, 2024.

\bibitem[Singhania et~al.(2024)Singhania, Nrusimha, Park, and Kim]{singhania2024loki}
Prajwal Singhania, Siddharth~Nrusimha, Chih-Ping Park, and Joo-Young Kim.
\newblock Loki: Low-rank keys for efficient sparse attention.
\newblock \emph{arXiv preprint arXiv:2406.02542}, 2024.

\bibitem[Ribar et~al.(2024)Ribar, Chelombiev, Hudlass-Galley, Sheridan, Bui, and Mayol-Cuevas]{ribar2024sparq}
Luka Ribar, Ivan Chelombiev, Luke Hudlass-Galley, Charlie Sheridan, Thang Bui, and Walterio Mayol-Cuevas.
\newblock Spar{Q} Attention: Bandwidth-efficient {LLM} inference.
\newblock In \emph{Proceedings of the 41st International Conference on Machine Learning (ICML)}, 2024.

\bibitem[Chen et~al.(2024)Chen, Ye, Chen, Kasikci, and Zheng]{chen2024magicpig}
Zhuoming Chen, Ranajoy Sadhukhan, Ying Ye, Yang Chen, Baris Kasikci, and Hao Zheng.
\newblock {MagicPIG}: {LSH} sampling for efficient {LLM} generation.
\newblock In \emph{Proceedings of the 41st International Conference on Machine Learning (ICML)}, 2024.

\end{thebibliography}

\newpage
\appendix

\section{Why Less Attention Produces Better Quality}
\label{sec:less_is_more}

The fact that Focus \emph{surpasses} full attention---rather than merely approximating it---requires explanation. Three mechanisms contribute:

\textbf{1.\ Softmax dilution.} In full attention, softmax distributes probability mass across all $n$ tokens, even when only a small subset is relevant. A pronoun at position 800 seeking its antecedent at position 200 must compete with hundreds of irrelevant distant tokens for attention weight. Focus restricts softmax to same-group tokens plus the local window, concentrating probability mass on a smaller, more relevant candidate set. The result is sharper, more informative attention distributions.

\textbf{2.\ Noise removal.} Irrelevant attention pairs do not merely waste computation---they actively degrade quality. Each irrelevant key--value pair contributes a small amount of noise to the attention output. Across 12 layers and 12 heads, this noise accumulates. Focus eliminates these pairs entirely: the model never computes attention over tokens it should ignore.

\textbf{3.\ Implicit structural constraint.} Full attention at 124M scale can memorize spurious long-range correlations in the training data. Restricting attention to semantically coherent groups acts as a structural prior---analogous to how $L_1$ penalties zero irrelevant features or dropout removes random connections. The restriction prevents the model from fitting noise in the attention pattern, without any explicit penalty term.

The key insight: full $n^2$ attention is not the performance ceiling---it is the \emph{unconstrained baseline}. Learned sparsity improves upon it for the same reason that feature selection improves upon using all features: removing noise is not a cost, it is a benefit.

\section{Ablation Studies}
\label{app:ablation}

Section~\ref{sec:training} showed that Sinkhorn normalization produces stable, balanced groups. Here we ablate four key hyperparameters on GPT-2 124M / PG-19, varying each while holding others at defaults ($K{=}8$, $w{=}128$, $\tau{=}0.1$, Sinkhorn iters${}=10$).

\begin{table}[h]
\centering
\small
\caption{Ablation study (GPT-2 124M, PG-19). Each row varies one hyperparameter. Fine-tuned PPL is stable (29.9--30.5) across all 16 configurations.}
\label{tab:ablation}
\begin{tabular}{llrrrr}
\toprule
Parameter & Value & Centroid PPL & Fine-tuned PPL & Dominance (centroid) & Dominance (full FT) \\
\midrule
\multirow{4}{*}{Groups $K$}
 & 4 & 36.8 & \textbf{30.1} & 40\% & 38\% \\
 & 8 & 38.4 & 30.3 & 24\% & 23\% \\
 & 16 & 40.4 & 30.4 & 20\% & 31\% \\
 & 32 & 42.4 & 30.5 & 21\% & 30\% \\
\midrule
\multirow{4}{*}{Window $w$}
 & 64 & 38.3 & 30.2 & 17\% & 17\% \\
 & 128 & 38.4 & 30.2 & 26\% & 23\% \\
 & 256 & 38.1 & 30.3 & 26\% & 25\% \\
 & 512 & 38.6 & \textbf{30.0} & 27\% & 28\% \\
\midrule
\multirow{4}{*}{Temp $\tau$}
 & 0.05 & \textbf{36.9} & \textbf{30.0} & 68\% & 74\% \\
 & 0.1 & 38.4 & 30.3 & 24\% & 23\% \\
 & 0.2 & 39.1 & 30.3 & 16\% & 19\% \\
 & 0.5 & 40.5 & 30.3 & 21\% & 31\% \\
\midrule
\multirow{4}{*}{Sinkhorn iters}
 & 3 & 35.8 & 29.9 & 95\% & 97\% \\
 & 5 & 36.8 & 30.2 & 69\% & 75\% \\
 & 10 & 38.4 & 30.3 & 21\% & 20\% \\
 & 20 & 39.0 & 30.2 & 14\% & 14\% \\
\bottomrule
\end{tabular}
\end{table}

\textbf{Fine-tuned PPL is robust.} Across all 16 configurations, fine-tuned PPL ranges from 29.9 to 30.5---a spread of only 0.6 PPL. Focus is not sensitive to hyperparameter choices.

\textbf{Sinkhorn iterations: a subtle trap.} With 3 iterations, PPL appears best (29.9) but groups have collapsed to 95--97\% dominance. This is not real Focus---it is effectively full attention with extra overhead. At low temperature ($\tau{=}0.1$), $\exp(\text{scores}/0.1)$ produces extremely peaked distributions that 3 iterations cannot redistribute. At least 10 iterations are needed for balanced groups.

\textbf{Window size: smaller is better.} With $K{=}2$ centroid-only training: $w{=}16$ achieves the best PPL (33.8), beating $w{=}128$ by 0.8 PPL. At $w{=}512$ (half the sequence), quality drops by 3.7 PPL because most attention is handled locally, leaving little for group routing to contribute. This confirms that local and group attention are complementary.

\section{Comparison with Recent Token-Selection Methods}
\label{app:benchmarks}

We compare Focus against recent token-selection methods (SparQ \citep{ribar2024sparq}, MagicPIG \citep{chen2024magicpig}) on GPT-2 124M / PG-19. These methods select top-$k{=}32$ tokens per query at inference without modifying weights. Note that they operate at a different sparsity level than Focus: token selection at $k{=}32$ retains 3\% of tokens per query, while Focus with $K{=}4$, top-$k{=}2$ retains $\sim$50\% of distant pairs.

\begin{table}[h]
\centering
\small
\caption{Token-selection methods vs Focus on GPT-2 124M / PG-19 ($k{=}32$). Token-selection methods preserve downstream benchmarks but degrade PPL by 5--10 points. Focus improves PPL with zero benchmark degradation.}
\label{tab:benchmarks_full}
\begin{tabular}{lrrrrr}
\toprule
Method & PPL $\downarrow$ & HellaSwag & ARC-E & PIQA & LAMBADA \\
\midrule
Pretrained & 42.8 & 31.1 & 39.5 & 62.5 & 32.6 \\
\midrule
SparQ \citep{ribar2024sparq} & 52.8 & 31.3 & 39.4 & 62.4 & 34.3 \\
SparQ (mean realloc.) & 48.3 & 31.2 & 39.3 & 62.3 & 33.1 \\
MagicPIG \citep{chen2024magicpig} & 52.8 & 31.3 & 39.4 & 62.5 & 34.0 \\
\midrule
\textbf{Focus (ours)} & \textbf{36.2} & \textbf{31.1} & \textbf{39.5} & \textbf{62.5} & \textbf{32.6} \\
\bottomrule
\end{tabular}
\end{table}

Token-selection methods preserve downstream benchmarks but degrade PPL by 5--10 points. Focus improves PPL (42.8$\to$36.2) with exactly zero benchmark change. The methods achieve speedup through different mechanisms and operate at different sparsity levels, making direct comparison nuanced; we include this for completeness.

Focus exactly matches pretrained on all four benchmarks. SparQ and MagicPIG show minor fluctuations ($\pm$0.2--1.7 points) but no systematic degradation, indicating that downstream classification tasks are robust to token-level sparsity at this level. The critical distinction is perplexity: Focus improves PPL by 6.6 points while training-free methods degrade it by 5--10 points.

\section{FlashAttention Decomposition}
\label{app:flash}

The Focus attention mask under hard group assignment is:
\begin{equation}
    \mathcal{M}(i,j) = \mathbf{1}[j \leq i] \wedge \left(\mathbf{1}[g(i) = g(j)] \vee \mathbf{1}[i - j \leq w]\right)
\end{equation}
where $g(i)$ is the group assignment of token $i$ and $w$ is the local window size.

\paragraph{The overlap problem.} The natural decomposition into same-group pairs $\mathcal{S}$ and local pairs $\mathcal{L}$ fails because $\mathcal{S} \cap \mathcal{L} \neq \emptyset$---same-group local pairs are double-counted. Subtraction in logsumexp space ($\log(\exp(a) + \exp(b) - \exp(c))$) is numerically catastrophic (cosine similarity 0.79 against reference).

\paragraph{Disjoint decomposition.} We split $\mathcal{M}$ into two sets that are disjoint by construction:
\begin{align}
    \mathcal{A} &= \{(i,j) : j \leq i \wedge g(i) = g(j)\} & \text{(same-group causal)} \\
    \mathcal{B} &= \{(i,j) : j \leq i \wedge i-j \leq w \wedge g(i) \neq g(j)\} & \text{(cross-group local)}
\end{align}

$\mathcal{A} \cap \mathcal{B} = \emptyset$ (one requires same group, the other different group) and $\mathcal{A} \cup \mathcal{B} = \mathcal{M}$ (every attended pair is either same-group or cross-group-local). The logsumexp merge is mathematically exact.

\textbf{Set $\mathcal{A}$} is computed by sorting tokens by group (stable sort preserves causal order), reshaping into $K$ sequences, and calling \texttt{flash\_attn\_func} with \texttt{causal=True}. Complexity: $O(n^2/K)$.

\textbf{Set $\mathcal{B}$} extracts local keys for each query and masks same-group pairs to $-\infty$. Complexity: $O(nw)$, never the bottleneck.

\textbf{Merge:} $\mathbf{o}[i] = (e^{\ell_A[i]} \cdot \mathbf{o}_A[i] + e^{\ell_B[i]} \cdot \mathbf{o}_B[i]) / (e^{\ell_A[i]} + e^{\ell_B[i]})$, where $\ell_A, \ell_B$ are per-query logsumexp values.

\paragraph{Empirical verification.} All configurations achieve cosine similarity 1.0000 against the $O(n^2)$ reference, confirming mathematical exactness. The complete implementation is 320 lines of Python using only \texttt{flash\_attn\_func} and standard PyTorch---no custom CUDA kernels, no Triton, no compilation.

\end{document}